\documentclass[runningheads,a4paper]{llncs}

\usepackage{amssymb}
\usepackage{graphicx}
\usepackage{tikz}
\usepackage{algorithm2e}
\usepackage{pgfplots}
\pgfplotsset{compat=1.15}
\usepackage{booktabs}
\usepackage{multirow}
\usepackage{subcaption}
\usepackage{url}
\urldef{\mailsa}\path|{alfred.hofmann, ursula.barth, ingrid.haas, frank.holzwarth,|
\urldef{\mailsb}\path|anna.kramer, leonie.kunz, christine.reiss, nicole.sator,|
\urldef{\mailsc}\path|erika.siebert-cole, peter.strasser, lncs}@springer.com|    
\newcommand{\keywords}[1]{\par\addvspace\baselineskip
\noindent\keywordname\enspace\ignorespaces#1}

\begin{document}

\mainmatter  

\title{DecisiveNets:\\Training Deep Associative Memories \\to Solve Complex Machine Learning Problems}

\titlerunning{DecisiveNets}

%
%
\author{Vincent Gripon \and Carlos Lassance \and Ghouthi Boukli Hacene}%
%
\authorrunning{DecisiveNets}

\institute{IMT Atlantique and Mila}

%
%

\maketitle

\begin{abstract}
Learning deep representations to solve complex machine learning tasks has become the prominent trend in the past few years. Indeed, Deep Neural Networks are now the golden standard in domains as various as computer vision, natural language processing or even playing combinatorial games.  However, problematic limitations are hidden behind this surprising universal capability. Among other things, explainability of the decisions is a major concern, especially since deep neural networks are made up of a very large number of trainable parameters. Moreover, computational complexity can quickly become a problem, especially in contexts constrained by real time or limited resources. Therefore, understanding how information is stored and the impact this storage can have on the system remains a major and open issue. In this chapter, we introduce a method to transform deep neural network models into deep associative memories, with simpler, more explicable and less expensive operations. We show through experiments that these transformations can be done without penalty on predictive performance. The resulting deep associative memories are excellent candidates for artificial intelligence that is easier to theorize and manipulate.
\end{abstract}

\section{Introduction}

In recent years, Deep Neural Networks (DNNs) have experienced a succession of major breakthroughs that have gradually elevated them to the standard method in machine learning. 
DNNs obey the philosophy of replacing arbitrary human choices with automatic optimizations, directly derived from an objective function related to the problem to be solved.
Thus, once the contours of an architecture have been determined, its functionality is acquired through a complex process involving a large amount of data, random processes and various combinations aimed at obtaining an efficient and generalizable solution.

Contrary to many other methods, DNNs do not seem to suffer from a phenomenon of overfitting. Interestingly, the opposite seems to be true: larger architectures generalize better, for which there is still not a clear explanation (We refer the reader to~\cite{nakkiran2019deep} for a possible explanation). On the other hand, as the architectures (and their performance) grow, their complexity and explainability deteriorate. In areas where the decisions of DNNs need to be understood, or where the resources involved in the calculations are limited, a complex dilemma is thus faced that has given rise to an abundant literature (cf. Section~\ref{sec:rw}).

Instead of trying to explain these complex architectures, a promising strategy is to approximate trained models by simpler and less expensive ones. 
This is, for example, the interest of the knowledge distillation field~\cite{hinton2015distilling}.
But even if the predictive capacity is thus transmitted from complex models to simpler ones, the fact remains that the latter are difficult to explain and compute.
Why not try instead to approximate trained deep neural network architectures by models built on lighter and more explicable mechanisms, allowing an explicit account of how information is stored? In this chapter we provide elements in order to address such a question. 

While DNNs are constructed explicitly as mathematical functions, with the goal of associating inputs with outputs, other models take a different view, where neural networks become the receptacle of information elements that can be retrieved from a fraction of their content. In this vein, sparse associative memories are very simple, yet surprisingly powerful tools that can be deployed in budget-constrained systems.

In this chapter, we introduce a method for transforming a deep neural network model, optimized for a machine learning task, into a deep associative memory with similar performance, although using processes that are lighter to implement and simpler to visualize and interpret. We illustrate this capability using standardized datasets in the field of computer vision.

The outline of the chapter is as follows. In Section~\ref{sec:rw} we introduce related work. In Section~\ref{sec:background} we introduce notations, deep neural network and associative memories methodologies. In Section~\ref{sec:methodo}, we introduce our proposed method that aims at transforming deep neural networks into deep associative memories. In Section~\ref{sec:experiments} we perform experiments and discuss the obtained results, and finally we conclude in Section~\ref{sec:conclusion}.

\section{Related Work}
\label{sec:rw}

\subsection{DNN compression}

The deep neural network complexity has been a hot topic of interest for researchers in the past few years. There are many reasons to be interested in compression of DNNs, including lack of memory, constrained energy, maximum acceptable latency, or targeted data throughput.
However, many works~\cite{hacene2019processing} show that there exists a compromise between network size and accuracy, leading to enormous architectures to reach state-of-the-art performance in many domains including vision and natural language processing.

As an effort to reduce the size of DNNs while maintaining a high level of accuracy, some authors propose to rely on pruning, where neurons and/or weights can be removed during training based on a measure of their importance in the decision process. Pruning can be unstructured~\cite{han2015learning}, meaning that any set of weights and neurons can be removed. As a consequence, the resulting operators sparsity can be hard to leverage depending on the targeted hardware. On the contrary, other works aims at removing specific sets of neurons and weights which are more easily exploitable~\cite{liu2017learning}. To measure neurons and/or weights importance, numerous criteria have been introduced. For instance, in~\cite{li2016pruning}, the authors use the sum of absolute weights of each channel to select less important parameters. Soft Filter Pruning (SFP)~\cite{he2018soft} is another approach that dynamically prunes filters in a soft manner. In~\cite{yu2018nisp}, the authors propose the Neuron Importance Score Propagation (NISP), a method that estimates neurons scores using the reconstructed error of the last layer before classification when computing back-propagation. Yamamoto \textit{et al}~\cite{yamamoto2018pcas} use a channel-pruning technique based on an attention mechanism, where attention blocks are introduced into each layer and updated during training to evaluate the importance of each channel. In~\cite{ramakrishnan2020differentiable}, the authors propose a learnable differentiable mask, that aims at finding out during training process the less important neurons, channels or even layers and prune them. In~\cite{he2020learning}, the authors propose to give to the DNN the ability to decide during training which criterion should be considered for each layer when pruning. Another method that relates better to our proposed solution is Shift Attention Layers~\cite{hacene2019attention}. The idea is to prune all but one weight per convolutional kernel, which reduces both DNNs memory and complexity and eases DNNs implementation on limited resources embedded systems~\cite{hacene2018quantized}.

Another line of works consists in quantizing values and weights in DNNs, with the extreme case being binarizing both~\cite{hubara2016binarized,rastegari2016xnor}. As a consequence, memory requirements are heavily reduced and operations considerably simplified. Problematically, many such works end with a noticeable drop in accuracy of the corresponding architectures. In order to compensate for this drop, it is often required to increase the size of the architecture, leading to an unclear optimal trade-off in the general case. Other approaches propose to limit the accuracy drop by using a low bit quantization, where weights and/or activations are represented using $2$, $3$ or $4$ bits~\cite{lin2015neural,li2016ternary,zhou2016dorefa,esser2019learned}, or even learn the number of bits required to represent each layer values~\cite{nikolic2020bitpruning}. In such scenario, possible values are limited, quantized network size is reduced and multiplications may be replaced by Look Up Tables (LUTs)~\cite{wang2019lutnet}.

Other works propose to compensate the drop in accuracy by relying on knowledge distillation~\cite{hinton2015distilling}, where a bigger architecture (called the teacher) is used to train a smaller one (called the student).

The method we introduce in this chapter has yet a different approach, where instead of trying to reduce the complexity of a DNN architecture, we progressively transform it into a Deep Hetero-Associative Memory (DHAM), built from assembling lighter operators.

\subsection{Associative Memories}

Associative memories are devices able to store then retrieve pieces of information (called messages) from a portion of their content. The most prominent model is that proposed by John Hopfield in~\cite{hopfield1982neural}, where the principle consists in using the Gram matrix to store binary ($\{-1,1\}$) messages. Using a simple iterative algorithm, such messages can be reliably retrieved as long as their number remains small compared to the number of neurons in the architecture. In~\cite{mceliece1987capacity}, the authors proved that this bound evolves as $n/\log(n)$, contradicting the conjecture of many previous works that estimated this capacity to be linear with $n$.

To increase the number of messages it is possible to store, other authors have proposed to consider sparse binary ($\{0,1\}$) messages instead. Because the entropy each of these message contain is smaller, it is theoretically possible to store more of them. For example in~\cite{amari1989characteristics}, the authors do not update the retrieval iterative procedure proposed in~\cite{hopfield1982neural}. Better results are generally obtained using the model in~\cite{willshaw1969non}, where connection weights are thresholded at 1. More recently, in~\cite{gripon2011sparse}, the authors introduced a constraint on stored messages: they can be split into $c$ parts, where each part contains exactly one nonzero value. In~\cite{gripon2016comparative}, a comparative study of these models is performed, showing that the latter one can reliably store the largest number of messages. This number of messages evolves as $n^2 / \log(n)$, where $n$ is the number of neurons in the architecture.

In our work, we use the model introduced in~\cite{gripon2011sparse}, deployed in multiple layers to mimic the architecture of a DNN. We describe in details how to build such a model in the next section.

\section{Background}
\label{sec:background}

In this section we introduce the vocabulary and notations required for the remaining of this work. We begin with DNNs and continue with sparse associative memories.

\subsection{Deep Neural Networks}

Deep Neural Networks~\cite{lecun2015deep} (DNNs) are composite systems obtained by assembling \emph{layers}. Layers can be of various types and serve different purposes. In this work, we consider a generic model for a layer using a function: $f_l: \mathbf{x} \mapsto \sigma(\mathbf{W} \mathbf{x} + \mathbf{b})$, where $\mathbf{W}$ is a weight tensor, $\mathbf{b}$ is a bias tensor and $\sigma$ is a nonlinear activation function. Layers can be assembled using various combinators, including additions or concatenations. As a result, a composite function $f$ is obtained, which associates an input $\mathbf{x}$ with a corresponding decision $\hat{\mathbf{y}} = f(\mathbf{x})$.

We refer to $f$ as the \emph{network function}. Initially, the coefficients in the tensors $\mathbf{W}$ and $\mathbf{b}$ -- called \emph{parameters} -- are typically arbitrarily chosen, with no relation to the considered problem to be solved using the deep neural network.

During a learning phase, network functions are trained to solve a problem by optimizing an objective function on a training dataset. Most of the time, a training dataset is made of pairs $(\mathbf{x},\mathbf{y})$, where $\mathbf{y}$ is the ideal decision associated with $\mathbf{x}$. During the training phase, $\mathbf{y}$ and $\hat{\mathbf{y}} = f(\mathbf{x})$ are compared using an objective function. The derivative of this objective function is used to update the parameters of the deep neural network.

Once trained, the network function is expected to perform well on previously unseen data. This property, called the generalization, is often assessed using a validation set. A validation set is a dataset sampled from the same distribution as the training set, but disjoint from it. The prediction of the network function on the validation set is a measure of its ability to perform on inputs not used during training.

\subsection{Sparse Associative Memories}

We follow the model introduced in~\cite{gripon2011sparse}. Let us consider pairs $\left(\mathbf{x}_i, \mathbf{y}_i\right)_i$ that we want to store. Our aim is to learn the mapping $\mathbf{x}_i \mapsto \mathbf{y}_i$ using a neural network. We consider that vectors $\mathbf{x}_i$ and $\mathbf{y}_i$ have a specific shape. Namely, they are all binary ($\{0,1\}$) vectors. Vectors $\mathbf{x}_i$ are such that they can be split into $c$ subvectors, each containing $\ell$ consecutive coordinates. The resulting subvectors contain exactly one nonzero value. Similarly, vectors $\mathbf{y}_i$ can be split into $c'$ subvectors, each containing $\ell'$ consecutive coordinates. Each subvector contains exactly one nonzero value. An example of a vector $\mathbf{x}_i$, with $c=4$ and $\ell=3$ is given in Figure~\ref{fig:example_sparse}, together with an associated vector $\mathbf{y}_i$, with $c'=2$ and $\ell'=5$.

\begin{figure}
    \centering
    \begin{tikzpicture}
    \node at (-1,0) {$\mathbf{x}_i = $};
    \node() at (0,0) {$\left(\begin{array}{c}0\\0\\1 \\\hline 1\\0\\0 \\\hline 1\\0\\0 \\\hline 0\\1\\0 \end{array}\right)$};
    \node at (1,0) {$\mapsto$};
    \node at (2,0) {$\mathbf{y}_i = $};
    \node() at (3,0) {$\left(\begin{array}{c}0\\1\\0\\0\\0 \\\hline 0\\0\\0\\1\\0 \end{array}\right)$};
    \end{tikzpicture}
    \caption{Arbitrary example of a mapping between $\mathbf{x}_i$, built with $c=4$ and $\ell = 3$, and $\mathbf{y}_i$, built with $c'=2$ and $\ell'=5$.}
    \label{fig:example_sparse}
\end{figure}

In order to store such pairs, a matrix $\mathbf{W}$ is built using the following formula:
\begin{equation}
    \mathbf{W} = \max_i\left(\mathbf{y}_i \mathbf{x}_i^\top\right),
\end{equation}
where $\max$ is applied coefficient-wise, and $\mathbf{y}_i^\top$ is the transpose of $\mathbf{y}_i$. Let us point out that $\mathbf{W}$ is of dimensions $c'\ell' \times c\ell$.

Once $\mathbf{W}$ has been built, it can be used to try to retrieve $\mathbf{y}_i$ from $\mathbf{x}_i$. To this end, we use Algorithm~\ref{algo_retrieve}.

\begin{algorithm}
$\mathbf{z} = \mathbf{W} \mathbf{x}_i$\\
\For{$i \in \{1, \dots, c\}$}{
$\hat{\mathbf{y}_i}[\ell(i-1):\ell i] = \texttt{WTA}(\mathbf{z}[\ell(i-1):\ell i])$\\
}
\Return{$\hat{\mathbf{y}_i}$}\\
\caption{Algorithm to retrieve $\mathbf{y}_i$ from the corresponding input probe $\mathbf{x}_i$ and $\mathbf{W}$. We use Python notations here, where $\hat{\mathbf{y}_i}[\ell(i-1):\ell i]$ denotes the subvector obtained from $\hat{\mathbf{y}_i}$ by taking the coordinates $\ell(i-1)$ (included) to $\ell i$ (excluded). The operator $\texttt{WTA}$ is a binary winner-takes-all that outputs a 1 at coordinate $j$ if and only if it corresponds to the maximum value in the corresponding vector (0 otherwise).}
\label{algo_retrieve}
\end{algorithm}

In general, there is no guarantee that the retrieved vector $\hat{\mathbf{y}_i}$ is identical to $\mathbf{y}_i$. All depends on the  number of pairs used to generate $\mathbf{W}$ and to the potential overlap between the vectors in $\left(\mathbf{x}_i\right)_i$ or the vectors in $\left(\mathbf{y}_i\right)_i$. For a more in-depth analysis of this problem, we refer the reader to~\cite{GriHeuLoVer2016}.

Very much alike DNNs, it is possible to assemble associative memories to create deep architectures, called Deep Hetero-Associative Memories or DHAM in the following, where the output of one is the input of another. Problematically, associative memories are meant to store mappings between pairs of vectors. As such, there is very little interest in creating such assemblies, while the scope of possible applications remains very limited.

In the next section, we show how it is possible to create DHAMs able to solve complex machine learning problems, far beyond the case of storing mappings between pairs of vectors.

\section{Methodology}
\label{sec:methodo}

\subsection{Core principle}

Our proposed methodology is based on the following idea: it is possible to train a deep neural network, while progressively updating its operations so that it ultimately functions as a DHAM. Our motivation is threefold: 
\begin{enumerate}
    \item Because of the simplicity of the operations used in the context of DHAMs, we expect the resulting architecture to require significantly less operations to achieve a very similar performance to DNNs.
    \item As DHAMs operations are simple, it is quite straight-forward to understand the decisions taken by a specific layer, and to obtain guarantees about the decision process.
    \item Because DHAMs internal representations are very sparse, they are strong candidates for better transferability and robustness to deviations of the inputs.
\end{enumerate}

We shall investigate these points later in the experiments. But for now, let us explain how we can amend the training process of DNNs so that they ultimately function as DHAMs.

The main idea is to act on the nonlinear activation function $\sigma$ that is used at each layer of the considered DNN architecture. Starting with $\sigma$, the training process will smoothly and continuously transform it until it eventually becomes a local winner-takes-all operator. To this end, we make use of a temperature $t$, that is scaled exponentially from $t_{init}$ to $t_{final}$ at each step of the learning procedure. This temperature acts on a \texttt{softmax} function. More precisely, let us denote by $\mathbf{x}$ the input tensor of a layer, and by $\mathbf{\hat{y}}$ its output. We construct the following nonlinear function:

\begin{eqnarray*}
    \mathbf{\hat{y}}[\ell (i-1):\ell i] = \sigma_t\left(\mathbf{x}[\ell (i-1):\ell i]\right), \forall i, \texttt{ where}\\
    \sigma_{t}(\mathbf{z}) = \frac{\texttt{softmax}\left(t \cdot \sigma(\mathbf{z})\right)}{\max\left(\texttt{softmax}\left(t \cdot  \sigma(\mathbf{z})\right)\right)} \sigma(\mathbf{z}).
\end{eqnarray*}

Note that instead of $\ell$, it is possible to define such a function using $c$ as a parameter such that $c\ell$ is the total number of dimensions of the tensor $\mathbf{x}$ along the considered axis. In our experiments, we found that acting on the feature maps axis when using convolutional neural networks worked the best. We also found that it was best to disregard the gradients of the softmax portion of the activation.

At the beginning of the training process, the initial temperature is chosen so that it is close to 0. As a consequence the softmax operator outputs a constant vector and $\sigma_t = \sigma$. At the end of the training process, the temperature is very high, so that the softmax operator acts as a max operator.

Once the training process is finished, and before using our trained architecture for processing test inputs, we completely remove the softmax operator, and consider instead the following nonlinear function:

\begin{eqnarray*}
    \mathbf{\hat{y}}[\ell (i-1):\ell i] = \sigma_{WTA}\left(\mathbf{x}[\ell (i-1):\ell i]\right), \forall i, \texttt{ where}\\
    \sigma_{WTA}(\mathbf{z}) = \left(\texttt{max}\left(\sigma(\mathbf{z})\right) == \sigma(\mathbf{z})\right) \sigma(\mathbf{z}).
\end{eqnarray*}

This nonlinear function can be interpreted as the limit case of the previous one when the temperature tends to infinity.

\subsection{Interpretation}

In the case of convolutional neural networks, the resulting architecture thus implements a competition between feature maps. More precisely, inside a group of $\ell$ consecutive feature maps, each location (e.g. a pixel in the case of processing images) can activate only one feature map. When assembling the $c$ obtained subtensors, each location is thus summarized as a combination of the $c$ corresponding choices of activated feature maps. The number of possible combination is thus $\ell^c$ for each location. In our experiments, we found that choosing small values of $\ell$ and thus large values of $c$ usually led to the best results, which is in accordance with the fact that this choice maximizes the number $\ell^c$, and thus the diversity of possible values for each processed input.

Another important remark is that the resulting winner-takes-all function differs from that of DHAMs in the fact the remaining maximum value is kept as is, instead of being put to 1. We tested both possibilities in our experiments, with a slight advantage in keeping the value. We believe that the main reason for this difference lies in the fact that keeping the value allows for considering the case where a group of $\ell$ feature maps shows no clear winner.

In the limit case where $\ell = 1$, we retrieve the function $\sigma$, thus encompassing the untouched reference DNN architecture. On the contrary, when $\ell$ is maximum, we can only select one feature map for each location in the treated inputs. More generally, varying $\ell$ from 1 to its maximum value has the effect of reducing the number of possible combinations of feature maps, together with increasing sparsity and reducing computations. As a matter of fact, when processing the next layer, a larger value of $\ell$ causes more values to be nullified, and thus considerably reduces the number of multiplications to compute.

In the following section, we present our experiments and discuss the obtained results.

\section{Experiments}
\label{sec:experiments}

\subsection{Considered architectures and datasets}

For our experiments we make use of the CIFAR-10 and CIFAR-100 datasets. These two datasets are made for image classification purposes, and have the double advantage of being relatively small so that simulation time remains acceptable and being challenging enough to be considered for benchmarking purposes. CIFAR-10 comes with 10 classes and 5,000 examples per class for training, while CIFAR-100 comes with 100 classes and 500 examples per class for training. Both datasets are made of RGB images of 32x32 pixels.

Architecture-wise, we make use of Resnet-18~\cite{he2016deep} for CIFAR-10 and Resnet-50~\cite{he2016deep} for CIFAR-100. We motivate this choice by the fact that these architectures exhibit a very interesting trade-off between complexity and accuracy~\cite{hacene2019processing}. During training, we divide the learning rate by 10 twice, at epochs 100 and 200. We use a total of 300 epochs for training the architectures, with batches of size 128. We use standard data augmentation techniques comprising of random cropping and horizontal flipping. We report the average accuracy over 5 experiments for the CIFAR-10 experiments, while for the CIFAR-100 we only run one experiment due to timing constraints.

\subsection{Influence of $\ell$}

In a first series of experiments, we vary the parameter $\ell$. To considerably reduce the combinatorial space of possibilities in our architectures, we only considered the case of using the same value of $\ell$ for all layers, despite our belief that this might be suboptimal. We report the obtained results in Table~\ref{table:varying_ell}. We also report the number of multiplications required to process a single input. As expected, the method exhibits a trade-off between number of multiplications and accuracy.

We recall that $\ell=1$ corresponds to the untouched baseline DNN architecture. Interestingly, we observe that with both considered datasets, the proposed methodology was able to improve the accuracy of the system, while considerably reducing the number of multiplications.

\begin{table}[ht]
     \caption{Accuracy of considered architectures while varying $\ell$. The number of multiplications for processing an input is also reported.}
    \centering
    \begin{tabular}{|c|c|c|c|c|}
    \hline
    &\multicolumn{2}{c|}{Resnet18 and CIFAR-10}& \multicolumn{2}{c|}{Resnet50 and CIFAR-100}\\
    \hline
    $\ell$ & accuracy & multiplications & accuracy & multiplications \\
    \hline\hline
1&	95.21\%&	5070848&	78.50\%	&1297809408\\
2&	\textbf{95.25\%}&	3125248&	\textbf{79.23\%}	&861601792\\
4&	94.65\%&	2152448&	76.58\%	&643497984\\
8&	92.95\%&	1666048&	70.46\%	&534446080\\
16&	88.95\%&	1422848&	64.36\%	&479920128\\
32&	84.90\%&	1301248&	61.07\%	&452657152\\
64&	78.28\%&	1240448&	53.05\%	&439025664\\
\hline
    
    \end{tabular}
    \label{table:varying_ell}
\end{table}

\subsection{Influence of $c$}

Instead of varying $\ell$, it is possible to vary the parameter $c$. The difference is the following: in Resnet architectures, the number of feature maps is regularly scaled up while progressing deeper in the architecture. When $\ell$ is used, the effect is to increase the number of subvectors considered in deeper layers. When $c$ is increased, the effect is to increase the length of considered subvectors in deeper layers. The obtained results are presented in Table~\ref{table:varying_c}. Without surprise, the results we obtained using $c$ were less interesting than when using $\ell$. The reason is that when forcing $\ell=2$ we obtain the maximum number of possible combinations (that we cannot obtain while varying the $c$ parameter).

\begin{table}[ht]
     \caption{Accuracy of considered architectures while varying $c$.}
    \centering
    \begin{tabular}{|c|c|c|}
    \hline
    &Resnet18 and CIFAR-10& Resnet50 and CIFAR-100\\
    \hline\hline
1&	59.73\%	&		13.39\%\\
2&	70.69\%	&		26.34\%\\
4&	81.04\%	&		41.84\%\\
8&	87.38\%	&		55.27\%\\
16&	91.09\%	&		62.01\%\\
32&	94.00\%	&		65.89\%\\
64&	\textbf{94.91\%}	&		\textbf{71.87\%}\\
\hline
    
    \end{tabular}
    \label{table:varying_c}
\end{table}

\subsection{Influence of the initial and final temperatures}

In the previous experiments, we purposely did not report the choice of the initial and final temperatures. The results we indicated were obtained using a grid search for the best possible temperature parameters. In this subsection, we report typical results we found for this search. These are summarized in Table~\ref{table:varying_temp}. Interestingly, we observe very little dependence on the choice of these parameters, as long as the final temperature is high enough to ensure a smooth transition towards the winner-takes-all nonlinear activation function.

\begin{table}[ht]
     \caption{Accuracy of considered architectures while varying the initial and final temperatures and for various choices of $\ell$.}
    \centering
    \begin{tabular}{|c|c|c|c|c|}
    \hline
    $\ell$& $t_{init}$ & $t_{final}$ & Resnet18/CIFAR-10& Resnet50/CIFAR-100\\
    \hline\hline
    2&	0.1&	100&	94.56\%&	74.19\%	\\
	&1&	100&	95.01\%&			78.17\%	\\
	&10&	100&	95.24\%&		79.30\%	\\
	&0.1&	1000&	94.81\%&		77.93\%	\\
	&1&	1000&	95.25\%&			78.81\%	\\
	&10&	1000&	95.25\%&			79.23\%	\\
	&0.1&	10000&	95.07\%&			78.28\%	\\
	&1&	10000&	95.11\%&			78.89\%	\\
	&10&	10000&	95.13\%&			78.88\%	\\\hline
4	&0.1&	100&	90.35\%&	53.00\%	\\
	&1&	100&	93.65\%&		66.27\%	\\
	&10&	100&	94.65\%&		76.80\%	\\
	&0.1&	1000&	94.02\%&		74.61\%	\\
	&1&	1000&	94.72\%&		76.56\%	\\
	&10&	1000&	94.65\%&		76.58\%	\\
	&0.1&	10000&	94.49\%&		76.16\%	\\
	&1&	10000&	94.74\%&		76.73\%	\\
	&10&	10000&	94.73\%&		76.89\%	\\
\hline
    
    \end{tabular}
    \label{table:varying_temp}
\end{table}

\subsection{Transfer potential}

In the next experiment, we aim at assessing the potential of the proposed methodology for improving transfer performance. We use for this purpose the CIFAR-FS dataset, which is built from CIFAR-100 by splitting the classes into three groups. The first group made of 64 classes is considered for training a CNN backbone architecture (in our case transformed into a DHAM). A next group of 16 classes is meant for validation purposes; we disregard it in this experiment. The backbone architecture is then used to extract features from inputs belonging to 5 unseen classes chosen uniformly at random among the 20 remaining ones. We dispose of 5 examples per class to perform prediction, and evaluate the performance on the remaining 595 vectors per class. To do so, we rely on a simple nearest class mean classifier, known to reach top performance for this type of problem~\cite{wang2019simpleshot}. We perform an average over 10,000 runs obtained by varying the choice of the 5 classes among 20 and the choice of the 5 supervised examples per class. Results are summarized in Table~\ref{table:transfer}. Interestingly, with this experiment we can clearly that the performance obtained using DHAMs is significantly better than that with the considered baseline.

\begin{table}[ht]
     \caption{Accuracy of DHAMs when performing few-shot transfer classification on the CIFAR-FS dataset. The confidence interval at 95\% is indicated between parenthesis.}
    \centering
    \begin{tabular}{|c|c|c|}
    \hline
    $\ell$& Resnet18 & Resnet50\\
    \hline\hline
   1 & 82.74 ($\pm 0.13$) & 83.94 ($\pm	0.14$)\\
   2 & 83.78 ($\pm 0.14$) & \textbf{85.44} ($\pm 0.13$)\\
   4 & \textbf{84.21} ($\pm 0.13$) & 84.83 ($\pm 0.13$)\\
   8 & 81.87 ($\pm 0.14$) & 82.43 ($\pm 0.14$)\\
   16 & 78.63 ($\pm 0.15$) & 77.93 ($\pm 0.15$)\\
   32 & 72.09 ($\pm 0.16$) & 76.40 ($\pm 0.15$)\\
   64 & 63.69 ($\pm 0.17$) & 70.15 ($\pm 0.16$)\\
\hline
    
    \end{tabular}
    \label{table:transfer}
\end{table}

\subsection{Robustness towards input deviations}

As mentioned in our motivations, by replacing the usual operations of DNNs with the simple local winner-takes-all of DHAMs, we expect the architectures to better accommodate for small variations of their input. As a matter of fact, the selection of a maximum has the effect of removing a large portion of small contributions due to the noise, as long as they do not provoke a change in the winner selection. In the next experiment, we stress the ability of our proposed DHAMs to better accommodate for various additive noises. We consider the noises described in~\cite{hendrycks2019benchmarking}, that is to say: Gaussian, Shot and Impulse noises. They are considered with five levels of severity each. We report in Table~\ref{table:noise} the obtained average accuracy under this type of perturbations. We observe that DHAMs are able to considerably reduce the impact of noise, depending on the choice of $\ell$. Interestingly, larger values of $\ell$ can prove more robust to such deviations, despite loosing more on the clean test set performance.

\begin{table}[ht]
     \caption{Accuracy of DHAMs subject to various types of noise (architecture: Resnet18).}
    \centering
    \begin{tabular}{|c|c|c|c|c|}
    \hline
    $\ell$& clean data & Gaussian noise & Shot noise & Impulse noise\\
    \hline\hline
1&  95.21\% &	46.40\% &	59.50\% &	51.75\% \\
2&  \textbf{95.25\%} &	47.59\% &	60.21\% &	53.45\% \\
4&  94.65\% &	49.98\% &	61.99\% &	52.33\% \\
8&  92.95\% &	46.34\% &	58.07\% &	53.54\% \\
16& 88.95\% &	50.51\% &	60.26\% &	48.95\% \\
32& 84.90\% &	\textbf{56.56\%} &	\textbf{64.34\%} &	\textbf{54.78\%} \\
64& 78.28\% &	48.72\% &	55.60\% &	40.03\% \\
\hline
    
    \end{tabular}
    \label{table:noise}
\end{table}

\subsection{Visualization of the competition between feature maps}

As we mentioned in the previous section, the main interpretation of the proposed methodology is that it creates competition between feature maps (when the softmax is applied to this dimension of the processed tensors). In terms of entropy, it would be ideal for the gradient descent to make sure that each feature maps is used approximately as often as any other. In order to visualize this, we depict in Figure~\ref{fig:l=2} and Figure~\ref{fig:l=8} the proportion of wins of activations of feature maps of arbitrarily selected layers in a trained Resnet50 architecture when processing test images. Note that the proportion of wins of a given feature map represents the sum of the binary version of its pixels obtained using \texttt{WTA}. We observe that contrary to our expectations, there is a strong unbalance in the selection of feature maps. We believe feature maps that are highly selective are likely to better reflect class-specific features, resulting in better overall classification. In future work, it would be interesting to consider introducing feature map based pruning criteria based on how strong is this unbalance.

\begin{figure}
\centering

\begin{subfigure}[ht]{.98\linewidth}
\centering
\resizebox{\linewidth}{!}{
\begin{tikzpicture}
\begin{axis}[
width=\textwidth,
 height=6cm,
bar width=8pt,
xlabel={Feature maps},
ylabel style={align=center},
ylabel={Mean amount of wins in \\corresponding feature maps},
ymin=0,
xmin=0,
xmax=36,
 xtick={1.5,4.5,7.5,10.5,13.5,16.5,19.5,22.5,25.5,28.5,31.5,34.5},
xticklabels={$1$, $2$,$3$,$4$,$5$,$6$,$7$,$8$,$9$,$10$,$11$,$12$},
tickwidth=0pt]
\addplot[ybar,ybar legend, fill=blue] coordinates {(1,320.35) (4,991.77) (7,173.43) (10,988.74) (13,475.82) (16,53.97) (19,803.18) (22,914.49) (25,959.19) (28,162.56) (31,40.15) (34,376.51)};
\addlegendentry{Class 1}
\addplot[ybar,ybar legend,fill=red] coordinates {(2,183.07) (5,988.56) (8,160.2) (11,989.28) (14,471.77) (17,54.64) (20,830.01) (23,912.83) (26,915.32) (29,167.61) (32,45.19) (35,270.2)};
\addlegendentry{Class 2}
\end{axis}
\end{tikzpicture}
}
\caption{1st Layer}
\end{subfigure}

\begin{subfigure}[ht]{.98\linewidth}
\centering
\resizebox{\linewidth}{!}{
\begin{tikzpicture}
\begin{axis}[
width=\textwidth,
 height=6cm,
bar width=8pt,
xlabel={Feature maps},
ylabel style={align=center},
ylabel={Mean amount of wins in \\corresponding feature maps},
ymin=0,
xmin=0,
xmax=36,
 xtick={1.5,4.5,7.5,10.5,13.5,16.5,19.5,22.5,25.5,28.5,31.5,34.5},
xticklabels={$1$, $2$,$3$,$4$,$5$,$6$,$7$,$8$,$9$,$10$,$11$,$12$},
tickwidth=0pt]
\addplot[ybar,ybar legend, fill=blue] coordinates {(1,133.56) (4,114.71) (7,122.48) (10,163.28) (13,131.09) (16,136.88) (19,141.73) (22,153.53) (25,100.78) (28,186.57) (31,42.6) (34,18.33)};
\addlegendentry{Class 1}
\addplot[ybar,ybar legend,fill=red] coordinates {(2,125.89) (5,103.08) (8,145.65) (11,153.84) (14,137.09) (17,141.47) (20,147.08) (23,171.98) (26,104.22) (29,181.68) (32,36.5) (35,30.77)};
\addlegendentry{Class 2}
\end{axis}
\end{tikzpicture}
}
\caption{21st Layer}
\end{subfigure}

\begin{subfigure}[ht]{.98\linewidth}
\centering
\resizebox{\linewidth}{!}{
\begin{tikzpicture}
\begin{axis}[
width=\textwidth,
 height=6cm,
bar width=8pt,
xlabel={Feature maps},
ylabel style={align=center},
ylabel={Mean amount of wins in \\corresponding feature maps},
ymin=0,
xmin=0,
xmax=36,
 xtick={1.5,4.5,7.5,10.5,13.5,16.5,19.5,22.5,25.5,28.5,31.5,34.5},
xticklabels={$1$, $2$,$3$,$4$,$5$,$6$,$7$,$8$,$9$,$10$,$11$,$12$},
tickwidth=0pt]
\addplot[ybar,ybar legend, fill=blue] coordinates {(1,11.26) (4,1.17) (7,4.01) (10,11.19) (13,5.1) (16,9.61) (19,15.11) (22,12.24) (25,8.51) (28,3.53) (31,6.61) (34,3.57)};
\addlegendentry{Class 1}
\addplot[ybar,ybar legend,fill=red] coordinates {(2,4.67) (5,7.88) (8,11.68) (11,7.53) (14,13.14) (17,10.98) (20,11.16) (23,2.87) (26,5.92) (29,7.88) (32,12.35) (35,8.02)};
\addlegendentry{Class 2}
\end{axis}
\end{tikzpicture}
}
\caption{48th Layer}
\end{subfigure}

\caption{Amount of wins in the $12$ first feature maps of the [1st,21st,48th] layers of Resnet50 for classes 1 and 2 trained on CIFAR100 when $\ell=2$.}

\label{fig:classescomparison}
\end{figure}
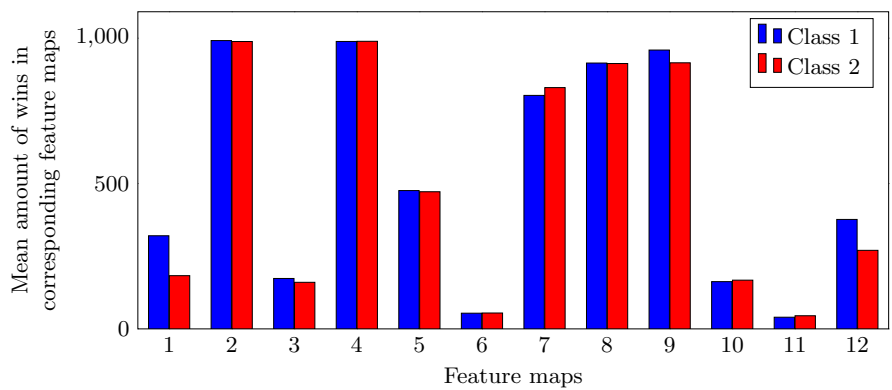
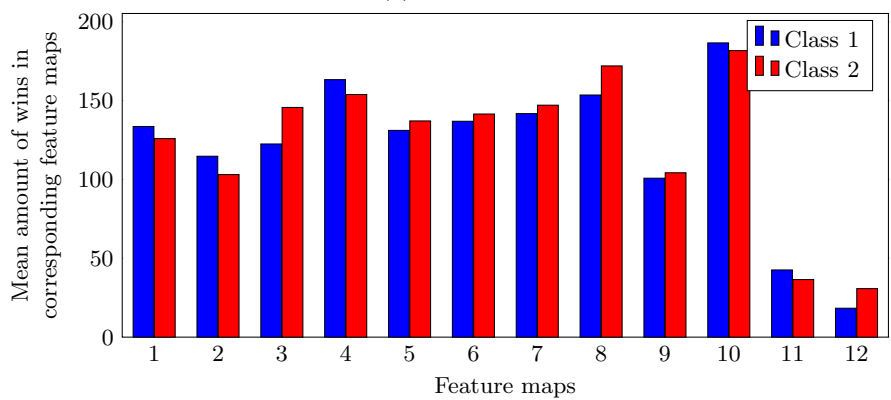
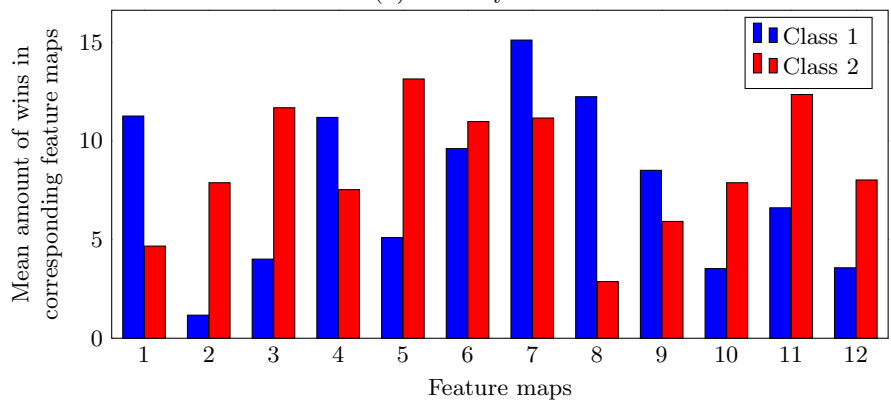

We perform a second experiment in which we aim at comparing the proportion of wins of features maps when processing inputs from different classes. To do so, we consider inputs from first and second class of CIFAR100 and report the proportion of wins of features maps of arbitrarily selected layer in a trained Resnet50. Figure~\ref{fig:classescomparison} compares the proportion of wins of the same feature map when considering two different classes, and shows that this proportion is almost the same in the first layer, a bit unbalanced in the 21st layer, and considerably unbalanced in the 48th (last) layer. Consequently, and as expected, first layers shown almost the same proportion of wins of feature maps since they are generic layers, while deeper layers shown an important difference of the proportion of wins since they are more specific.

\begin{figure}

    \centering
    
    \begin{tikzpicture}[scale=0.95]
    \begin{axis}[
    width=\textwidth,
    height=6cm,
      bar width=8pt,
      xlabel={subvectors},
      ylabel style={align=center},
      ylabel={proportion of wins in\\corresponding subvectors (\%)},
      ymin=0,
      xmin=0,
      xmax=36,
      xtick={1.5,4.5,7.5,10.5,13.5,16.5,19.5,22.5,25.5,28.5,31.5,34.5},
       xticklabels={$1$, $2$,$3$,$4$,$5$,$6$,$7$,$8$,$9$,$10$,$11$,$12$},
       tickwidth=0pt
    ]
      \addplot[ybar,fill=blue] coordinates {
 (1,16.6015625)
(2,83.3984375)
};

      \addplot[ybar,fill=green] coordinates {
(4,96.875)
(5,3.41796875)};
\addplot[ybar,fill=white] coordinates {
(7,15.234375)
(8,93.5546875)};
\addplot[ybar,fill=yellow] coordinates {
(10,96.09375)
(11,13.671875)};
\addplot[ybar,fill=brown] coordinates {
(13,47.94921875)
(14,52.1484375)};
\addplot[ybar,fill=black] coordinates {
(16,4.78515625)
(17,95.21484375)};
\addplot[ybar,fill=purple] coordinates {
(19,79.78515625)
(20,20.80078125)};
\addplot[ybar,fill=pink] coordinates {
(22,87.890625)
(23,34.27734375)};
\addplot[ybar,fill=orange] coordinates {
(25,88.8671875)
(26,15.625)};
\addplot[ybar,fill=blue] coordinates {
(28,13.76953125)
(29,88.28125)};
\addplot[ybar,fill=cyan] coordinates {
(31,4.19921875)
(32,96.09375)};
\addplot[ybar,fill=lime] coordinates {
(34,19.3359375)
(35,85.546875)};

    \end{axis}
  \end{tikzpicture}
    \caption{Proportion of wins for classes 1 and 2 in the $12$ first subvectors of the first layer of Resnet50 trained on CIFAR100 when $\ell=2$.}
    \label{fig:l=2}
\end{figure}
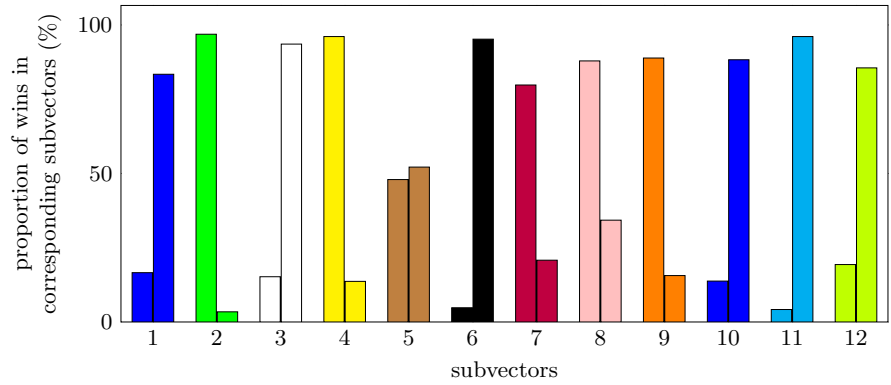

\begin{figure}
    \centering
    \begin{tikzpicture}[scale=0.95]
    \begin{axis}[
       width=\textwidth,
    height=6cm,
      bar width=8pt,
      xlabel={subvectors},
      ylabel style={align=center},
      ylabel={proportion of wins in\\corresponding subvectors (\%)},
      ymin=0,
      xmin=0,
      xmax=35,
      xtick={5.5,14.5,23.5,32.5},
       xticklabels={$1$, $2$,$3$,$4$},
       tickwidth=0pt
    ]
      \addplot [ybar,fill=blue] coordinates {
      
(1,0.0)
(2,9.765625)
(3,1.85546875)
(4,0.0)
(5,0.0)
(6,0.0)
(7,88.28125)
(8,0.0)
};
\addplot [ybar,fill=green] coordinates {
(10,2.44140625)
(11,0.0)
(12,6.54296875)
(13,0.0)
(14,16.89453125)
(15,25.29296875)
(16,48.53515625)
(17,0.0)};
\addplot [ybar,fill=brown] coordinates {
(19,6.25)
(20,0.0)
(21,59.9609375)
(22,0.0)
(23,0.0)
(24,0.29296875)
(25,33.30078125)
(26,0.0)};

\addplot [ybar,fill=purple] coordinates {
(28,5.078125)
(29,23.046875)
(30,61.23046875)
(31,0.0)
(32,2.34375)
(33,8.0078125)
(34,0.0)
(35,0.0)};
     
    \end{axis}
  \end{tikzpicture}
    \caption{Proportion of wins in the $4$ first subvectors of the first layer of Resnet50 trained on CIFAR100 when $\ell=8$.}
    \label{fig:l=8}
\end{figure}
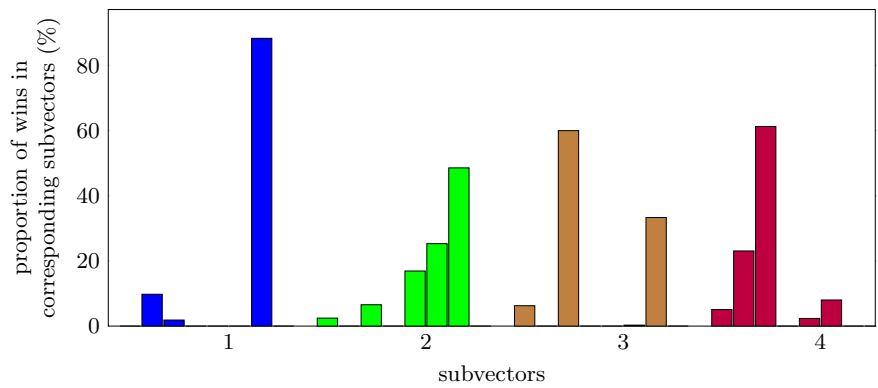

\section{Conclusion}
\label{sec:conclusion}

In this chapter we introduced a methodology to progressively transform classical deep neural networks, able to solve complex tasks such as classification of images, into deep hetero-associative memories, with much lighter and simpler operators. We show via experiments that this conversion can be performed with no cost on accuracy, while drastically reducing the number of multiplication. They also show that we are able to improve the performance in transfer learning and the robustness towards deviations of the inputs.

The proposed methodology allows to train sparse hetero-associative memories to solve any complex problem that can be solved using DNNs, despite the nondifferentiability of its main operators. For this reason, we believe that it is a promising direction of research for expanding the potentiality of these systems, even in very competitive engineering domains.

In future work, we would like to better understand the decisions taken by DHAMs using visualization tools. We also think it would be desirable to implement such systems on specific hardware, demonstrating the full potential of this solution.

\bibliography{bib}
\bibliographystyle{IEEEtran}
\end{document}